\documentclass[letterpaper, 10 pt, conference]{ieeeconf}  

\IEEEoverridecommandlockouts                              

\usepackage{graphics} 
\usepackage{epsfig} 
\usepackage{mathptmx} 
\usepackage{times} 
\usepackage{amsmath} 
\usepackage{amssymb}  
\usepackage{subfigure}
\usepackage{multirow}
\usepackage{graphicx, epstopdf}
\usepackage{xspace,epsfig,url}
\usepackage[noadjust]{cite}
\usepackage{enumerate}
\usepackage{float}
\usepackage{epsf}
\usepackage{psfrag}
\usepackage{verbatim}
\usepackage[all]{xy}
\usepackage{color}
\usepackage[table]{xcolor}
\usepackage{bm}
\usepackage{cases}
\usepackage{hyperref}
\usepackage[ruled,vlined,linesnumbered]{algorithm2e}
\usepackage{amssymb}
\usepackage{pifont}

\usepackage{algorithm2e}
\SetAlFnt{\footnotesize}
\SetAlCapFnt{\footnotesize}
\SetAlCapNameFnt{\footnotesize}
\usepackage{algorithmic}
\algsetup{linenosize=\tiny}

\newcommand{\CommaPunct}{\mathpunct{\raisebox{0.0ex}{,}}}

\title{\LARGE \bf
The Impact of Evolutionary Computation on Robotic Design: \\ A Case Study with an Underactuated Hand Exoskeleton
}

\author{Baris Akbas$^{1}$, Huseyin Taner Yuksel$^{1}$, 
Aleyna Soylemez$^{1}$, Mazhar Eid Zyada$^{2}$,\\Mine Sarac$^{2}$,~\IEEEmembership{Member,~IEEE}, and Fabio Stroppa$^{1}$,~\IEEEmembership{Member,~IEEE}
\thanks{This work is funded by TUB\.ITAK 
project number 123M690 and partially funded by TUB\.ITAK 
project number 121C145 and 121C147.}
\thanks{$^{1}$Computer Engineering, Kadir Has University, \.Istanbul, Turkey. E-mail: {\tt\small akbassbars99@stu.khas.edu.tr}}%
\thanks{$^{2}$Mechatronics Engineering, Kadir Has University, \.Istanbul, Turkey.}
}

\begin{document}

\maketitle
\thispagestyle{empty}
\pagestyle{empty}

\begin{abstract}
Robotic exoskeletons can enhance human strength and aid people with physical disabilities. However, designing them to ensure safety and optimal performance presents significant challenges. Developing exoskeletons should incorporate specific optimization algorithms to find the best design. This study investigates the potential of Evolutionary Computation (EC) methods in robotic design optimization, with an underactuated hand exoskeleton (U-HEx) used as a case study. We propose improving the performance and usability of the U-HEx design, which was initially optimized using a naive brute-force approach, by integrating EC techniques such as Genetic Algorithm and Big Bang-Big Crunch Algorithm. Comparative analysis revealed that EC methods consistently yield more precise and optimal solutions than brute force in a significantly shorter time. 
This allowed us to improve the optimization by increasing the number of variables in the design, which was impossible with naive methods. The results show significant improvements in terms of the torque magnitude the device transfers to the user, enhancing its efficiency. These findings underline the importance of performing proper optimization while designing exoskeletons, as well as providing a significant improvement to this specific robotic design. 
\end{abstract}


\section{Introduction}


Exoskeleton robotic devices are often used to augment users' strength and endurance during physically demanding tasks~\cite{poliero2020applicability,mauri2019mechanical,butzer2021fully}, to allow users control a secondary robotic device during teleoperation scenarios~\cite{koyama2002multi}, or to aid limited movements for patients with neurological and physical disabilities~\cite{stroppa2017robot}. Depending on the application, such exoskeletons can be designed for the whole body~\cite{marcheschi2011body} or for specific body locations such as arms~\cite{gijbels2011armeo}, legs~\cite{zoss2006biomechanical}
, wrists~\cite{buongiorno2018wres}, or hands~\cite{sarac2017design}. 
Regardless of the application or the body location, exoskeletons are very challenging to be designed, implemented, and controlled~\cite{sarac2019design,stroppa2023optimizing}. Safety is the primary and most important issue: exoskeleton joints (i) must align perfectly with anatomical joints to avoid potential harm, (ii) should work effectively within the workspace of human anatomical joints, and (iii) should allow the exoskeleton to follow users' behavior without creating discomfort. Finally, these devices –- especially assistive ones –- should be as compact and lightweight as possible to enhance wearability. 


\begin{figure}[b!]
    \centering
    \includegraphics[width=7.5cm]{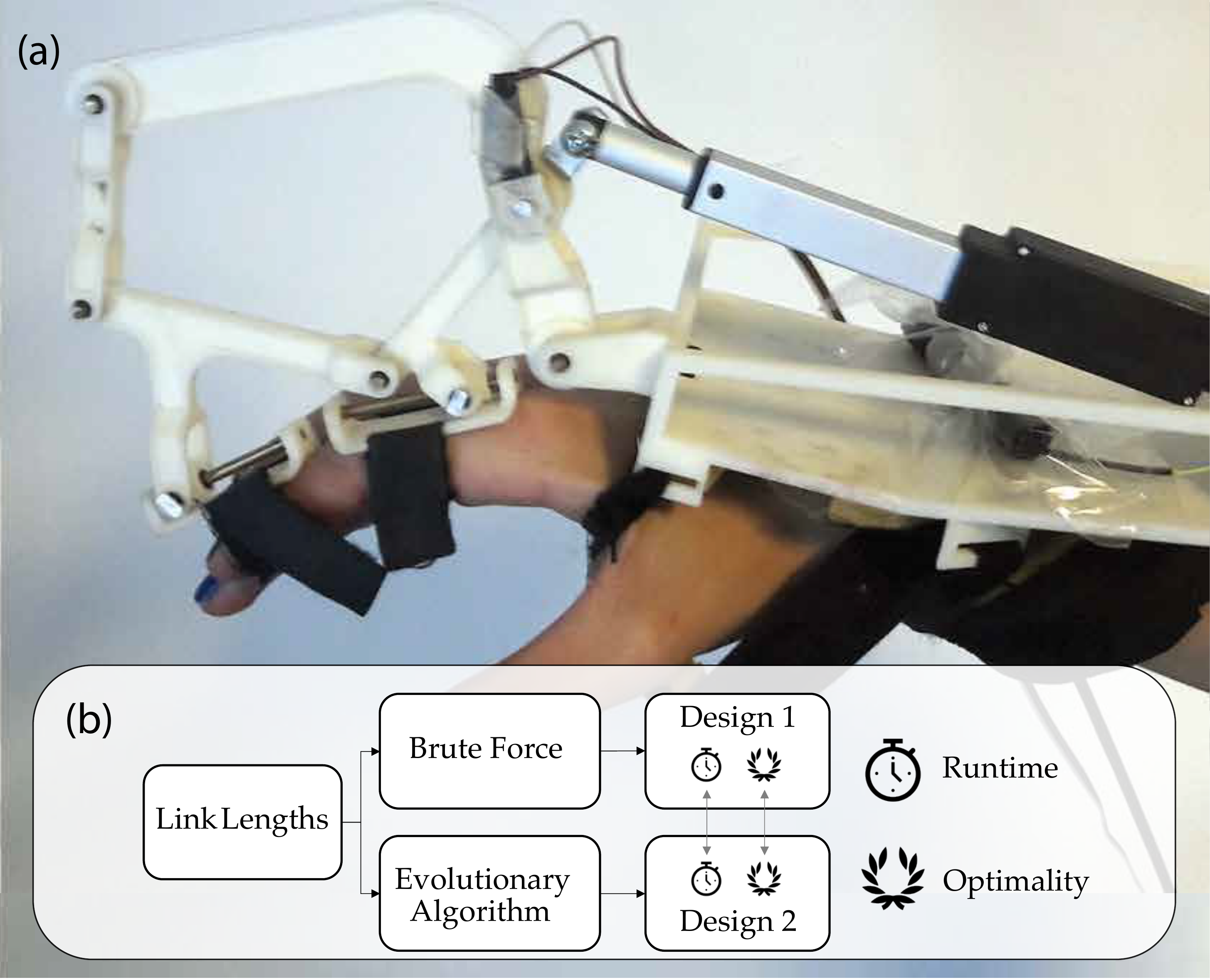}
    \caption{U-HEx: (a) A user wearing the first prototype of U-HEx from the state-of-the-art, which is bulky and cumbersome~\cite{sarac2017design}. Note that the picture only shows the device with a single finger. (b) Schema of the comparison between the approaches in terms of run time and optimality of the solution.}
    \label{fig:teaser}
\end{figure}


These challenges can be overcome by designing exoskeletons that can reach high output forces and feature effective power transmission despite using small, lightweight actuators. 
Thus, the design process of such exoskeletons should be integrated with various optimization algorithms; that is, the search for the best element within a set of alternatives based on specific criteria. Optimization is a common tool for solving engineering problems~\cite{sioshansi2017optimization, statnikov2012multicriteria, andersson2000survey}. While the most conventional strategies focus on numerical and calculus-based methods~\cite{bonnans2006numerical}, they might not be the best solution for engineering designs due to their properties such as non-discrete domains, non-differentiability, multi-modality, discontinuity, reliability, and robustness. 
Alternatively, the nature-inspired methods of Evolutionary Computation (EC) appear to be a common and effective way to deal with engineering optimization problems~\cite{dumitrescu2000evolutionary} -- and often with exoskeleton design~\cite{stroppa2023optimizing}.
Unfortunately, the integration between design and its optimization is not always straightforward. Roboticists, who often have mechanical/mechatronics backgrounds, might not know the latest trends or trade-offs in optimization. This is even more exacerbated by the lack of systematic studies in the literature on the impact of different optimization techniques for robotic design.


In this work, we attempt to fill the gap in the literature by raising awareness in the robotic community while displaying the impact of EC on robotic designs. We provide a systematical analysis of its performance on a \textit{poorly optimized} design from the literature~\cite{sarac2017design}. Compared to a naive-deterministic search approach (i.e., iteratively exploring each possible combination of design parameters -- ``brute-force''), we hypothesize that EC techniques \textit{(i)} might converge to a more precise and better solution (\textbf{H1}) and \textit{(ii)} will reduce the convergence time despite exploring border search spaces (\textbf{H2}). In addition, using various EC algorithms might offer different solutions at different convergence times \textbf{(H3)}.


Specifically, we applied the brute-force optimization approach and two EC algorithms on the design of an \textit{Underactuated Hand Exoskeleton} (U-HEx~\cite{sarac2017design}, shown in Fig.~\ref{fig:teaser} (a)), firstly on the same (limited) search space already explored in previous works, and then on a wider search space -- which could not be achieved with brute force. We specifically chose U-HEx due to its complex mechanism. Unlike conventional serial-link robots, its interconnected kinematics model makes it challenging to predict the contribution of each link length to the achieved range of motion.
We believe that this complexity highlights the differences between optimization methods and helps us formulate a clearer and better discussion. The comparison of the final designs will analyze both the optimality of the solution (i.e., effective force transmission) and the run time, as shown in Fig.~\ref{fig:teaser} (b).


\section{Background}
\label{sec:background}

\subsection{The Underactuated Hand Exoskeleton (U-HEx)}
\label{sec:uhex_background}


Fig.~\ref{fig:teaser} (a) shows U-HEx -- a wearable robotic device for the hand to rehabilitate stroke survivors through physical therapy~\cite{sarac2017design}. 
U-HEx is designed with a single actuator to control two finger joints through underactuation~\cite{laliberte2002underactuation}. With no external forces on the phalanges, the actuator opens and closes the finger naturally. As the user interacts with physical objects, the underactuated mechanism modifies the transmitted forces to each finger joint. Ultimately, U-HEx can automatically adjust its behavior based on the interaction forces -- allowing users to grasp objects with different shapes and sizes using a single actuator with no prior mechanical or control adjustments~\cite{buryanov2010proportions}. 

Furthermore, U-HEx promotes enhanced safety in multiple aspects of human-robot interaction. Firstly, the finger phalanges are considered as a part of the kinematic chain
-- such that the device is self-adaptable to a predefined range of hand sizes. Secondly, the underactuated kinematics inherently decouples the mechanical joints from the anatomical ones, so there is no need for calibration. The forces acting on the finger joints are transferred through the complicated design of mechanical links of U-HEx from the single actuator. Therefore, deciding the mechanical link lengths is of the utmost importance to ensure effective force transmission despite its complex kinematics chain. 


\subsection{Optimization and Evolutionary Computation}
\label{sec:optimization_background}

Optimization is the mathematical process of searching for a set of decision variables to minimize or maximize one or more specific objective functions while satisfying certain constraints. Optimization problems can be solved using methods that systematically and efficiently create and compare solutions to find the best outcome --  namely, \textit{Local Search} techniques. These methods can be exact, providing the precise optimal solution based on direct or gradient-based approaches~\cite{bonnans2006numerical}, or they can be approximate, yielding sub-optimal solutions that are acceptable approximations of the global optimum. In engineering design problems, approximate methods are often favored due to the complexity of objective functions~\cite{norde2000characterizing}, the preference for robust and reliable solutions over global ones~\cite{nomaguchi2016robust}, and the presence of uncertainty in the search space~\cite{dizangian2015reliability}. 

Evolutionary Computation (EC), a sub-field of soft computing, offers popular approximate methods for engineering~\cite{stroppa2023optimizing}. Inspired by natural selection, EC techniques generate a population of potential solutions and evolve them toward the optimal solution using different metaheuristics (i.e., based on genetic recombination, a storm of birds foraging for food, a community of bugs building a colony, etc.). These population-based techniques allow for parallelized search and the retrieval of multiple optimal solutions, particularly in multi-modal or conflicting multi-objective problems. 
Engineers find EC methods beneficial because they can effectively deal with objective functions defined implicitly through simulations and not rely on a specific mathematical model. This flexibility allows the same algorithmic implementation to be applied to different problems.

\section{Exoskeleton Design Optimization}
\label{sec:methods}

\subsection{U-HEx Kinematics}  

\begin{figure}[t!]
    \centering
    \includegraphics[width=8.5cm]{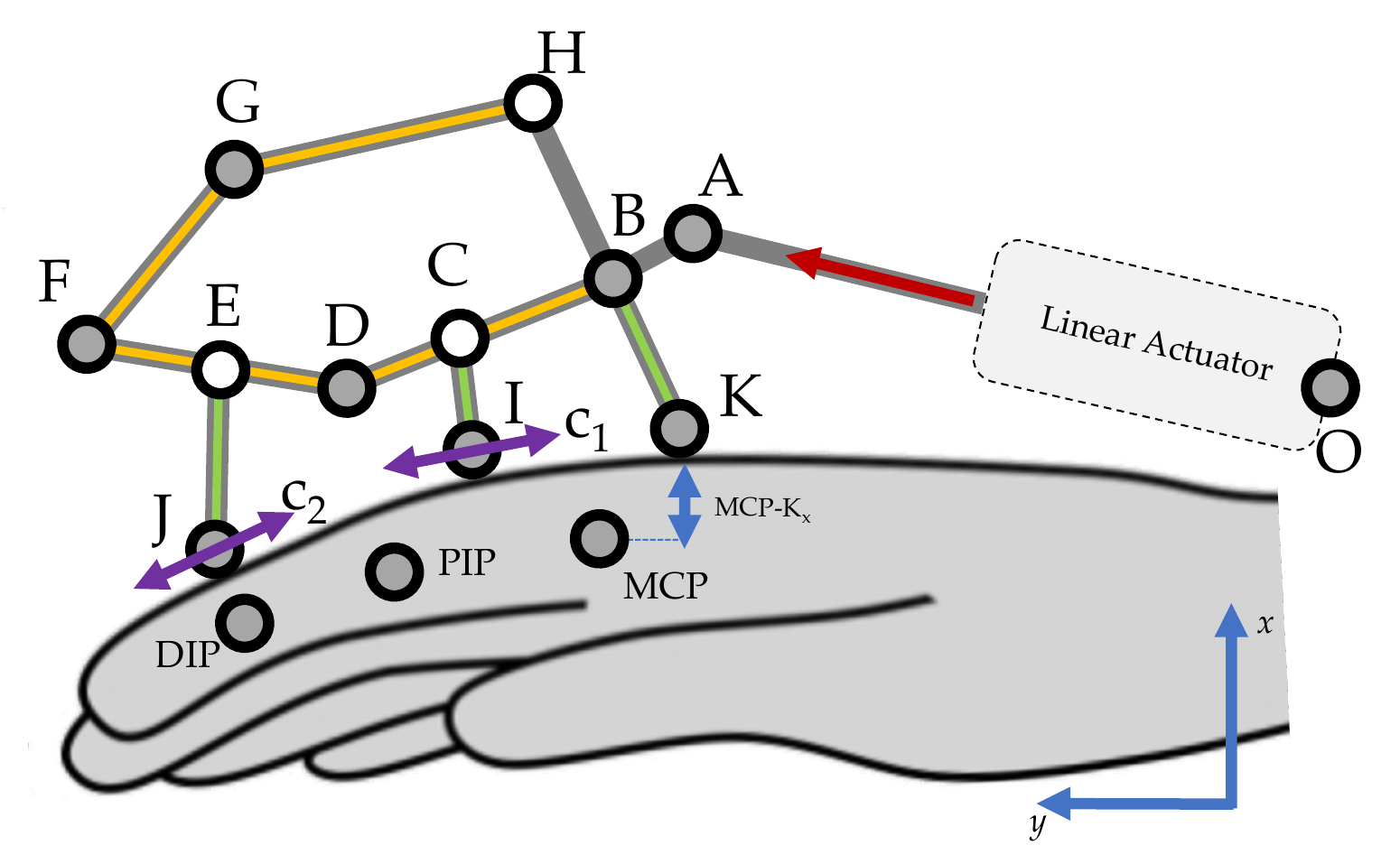}
    \caption{Kinematic model of U-HEx depicting all link lengths, passive joints (rotational and linear), and the active linear actuator. The colored links represent the decision variables of the optimization problem (i.e., the links to be optimized): in yellow, the ones that were used in its original design~\cite{sarac2017design}; in green, the additional three that were included in the current work thanks to evolutionary computation. The rest of the links are set to fixed values to ensure the kinematic chain is closed. Only joint $O$ is actuated.}
    \label{fig:u-hexkinem}
\end{figure}

Fig.~\ref{fig:u-hexkinem} shows the kinematic model for U-HEx with important points depicted with letters as detailed in previous works \cite{sarac2017design}. Each gray dot represents a passive rotational joint (or anatomical finger joints) while empty circles ($H$, $E$, and $C$) are fixed points along rigid links with $90^o$ fixed angle. The system has only one linear actuator between the points $O$ and $A$. The exoskeleton is fixed on the hand from points $K$ and $O$ with variable lengths in the $x$ and $y$ directions. Finally, the exoskeleton is attached to the finger phalanges from points $I$ and $J$ -- with passive linear joint sliders represented as $c_1$ and $c_2$, respectively. 

Defining the closed-loop kinematic chains with the letters depicted in Fig.~\ref{fig:u-hexkinem}, inverse kinematics are computed via numerical methods to compute $8$ unknown variables ($l_{OA}$, $q_O$, $q_A$, $q_B$, $q_G$, $q_D$, $c_1$, $c_2$) for given finger pose ($q_{MCP}, q_{PIP}$ -- from fully open to fully closed) and the set of link lengths. Once the kinematics are computed, the distribution from the unit actuator forces (1 N) to the torques around the finger joints ($\tau_{MCP}, \tau_{PIP}$) can be obtained either through static equations or the Jacobian of the system. 

\subsection{Link-Length Optimization Problem}  
\label{sec:link_length_opt}

The mechanical design of U-HEx should be optimized by searching for the set of link lengths (decision variables) that maximizes its force transmission (objective function) -- i.e., the amount of force transferred to the finger joints ($\tau_{i}$), as formulated in Eqn. (\ref{eq_op}). The problem is also subjected to the following physical constraints: \textit{(i)} U-HEx is connected to the user’s fingers through passive linear sliders ($c1$ and $c2$ in Fig.~\ref{fig:u-hexkinem}), whose movements are limited by the user’s finger size 
and \textit{(ii)} the ratio between the torques exerted on two finger joints must be within a reasonable range at different orientations of the finger (between $0.05$ and $20$).

\begin{equation}
    \label{eq_op}
    \begin{aligned}
    \textrm{maximize}  \quad & \sqrt{\tau_{MCP} + \tau_{PIP}} \\
    \end{aligned}
\end{equation}

While the hand exoskeleton has many link lengths that needed to be set, the previous study only optimized the important set of link lengths that were found through sensitivity analysis \cite{sarac2017design}. These important link lengths are depicted with yellow-lined links in Fig.~\ref{fig:u-hexkinem} (i.e., $\overline{BC}$, $\overline{CD}$, $\overline{DE}$, $\overline{EF}$, $\overline{FG}$, and $\overline{GH}$). Green-lined links indicate the additional decision variables that were included in the current study thanks to EC ($\overline{BK}$, $\overline{CI}$, and $\overline{EJ}$). Gray-lined links indicate the link lengths that are always kept constant.  

All optimization methods are performed using MATLAB script. For a decided set of link lengths to be tried, a Simulink model is executed with a fixed-step solver to compute \textit{(i)} the inverse kinematics through numerical methods and \textit{(ii)} the statics through analytical methods as the finger joints are iterated from fully open (0 deg each) to fully closed (80 deg for MCP and 90 deg for PIP). Once the Simulink file is terminated, we first check the constraints on the passive sliders ($c_1$ and $c_2$) 
and then the ratio between the transmitted torques ($\frac{\tau_{MCP}}{\tau_{PIP}}$). If the given set of link lengths satisfies these constraints, the objective function is computed for the finger pose fully closed.

\subsection{Optimization Methods}
\label{sec:methods}

\subsubsection{Brute Force (BF)}
\label{sec:brute_force}

The optimization method originally implemented to design U-HEx aimed at evaluating every solution in the search space~\cite{sarac2017design}. This is a brute-force method to solve an optimization problem; therefore, it is highly inefficient regarding both time and computational resources. Since the decision variables are lengths (i.e., measured in millimeters), their domain is considered almost continuous, making the search space too wide (theoretically infinite) for being treated with brute force -- even when the variables are bounded to specific lower and upper limits. Due to the enormity of this search space, the designers were forced to introduce the following limitations:

\begin{itemize}
    \item discretize the continuous domain of the decision variables by sampling with a fixed interval/step;
    \item increase the step between two contiguous discretized values for each decision variable (i.e., reduce the precision in millimeters); and
    \item reduce the number of decision variables, fixing the values of specific link lengths -- which were selected through sensitivity analysis to identify the ones that do not significantly affect the output performance.
\end{itemize}

The time complexity of such an algorithm is $\mathcal{O}(n^d)$, where $d$ is the number of decision variables, and $n$ is the cardinality of their domain (assumed to be the same for each variable or equal to the variable with the highest cardinality). When U-HEx was originally designed, this execution took approximately three days, during which the system crashed several times due to excessive processing (tested on a 2014 machine with a 2.40 GHz CPU and 16 GB RAM).

\subsubsection{Genetic Algorithms (GAs)}

GAs are the most popular methods of EC techniques as they directly implement the process of natural selection and survival of the fittest~\cite{goldberg1989genetic,goldberg1990real}. About Algorithm~\ref{alg:genetic}, they \textit{(i)} generate a population of random solutions $P$ within the search space of the problem, 
\textit{(ii)} assign a fitness value to each solution based on the objective function, and \textit{(iii)} generate new solutions $Q$ by mixing the values of the ones in the current population (i.e., a process named crossover). 

By allowing only the most-fitting solutions $M$ to perform crossover and be preserved in the next generations, GAs evolve their population to converge to the optimum of the problem. Like in biology, the operation of crossover exploits the features of good solutions (parents) to produce similar new solutions (offspring) and speed up convergence to an optimum; however, there is no guarantee for the optimum to be global rather than local. Therefore, GAs implement an additional operator inspired by genetic mutation, randomly modifying values of a newly generated solution to favor search-space exploration and escape local optima. 

Besides the common advantages of population-based methods (see Sec.~\ref{sec:optimization_background}), GAs are easy to implement and very efficient to converge. On the other hand, 
their many genetic operators come with many parameters and different types, and their fine-tuning might be non-trivial and primarily based on trial and error. Furthermore, since GAs are iterative stochastic methods, they might be inefficient for real-time problem solving -- which is not the case for this study.

\IncMargin{1em}
\begin{algorithm}[t!]
        \SetKwData{P}{P}
        \SetKwData{X}{M}
        \SetKwData{Q}{Q}
        \SetKwData{G}{g}
        \SetKwData{F}{f}
        \SetKwData{N}{n}
	\SetKwFunction{Init}{randomInitialization}
        \SetKwFunction{Eval}{evaluation}
        \SetKwFunction{Sel}{selection}
        \SetKwFunction{Var}{variation}
        \SetKwFunction{Sur}{survival}
	\SetKwInOut{Input}{input}
	\SetKwInOut{Output}{output}
        \Input{Population size $\N$, number of generations $\G$}
	\Output{The most fitting solution $\P(1)$}
	\BlankLine		
	
	\Begin{	

    	$\P \leftarrow \Init(\N)$\;
            $\P \leftarrow \Eval(\P)$\;
            \For{$i \in [1, \G]$}
            {
                $\X \leftarrow \Sel(\P)$\;
                $\Q \leftarrow \Var(\X)$\;
                $\Q \leftarrow \Eval(\Q)$\;
                $\P \leftarrow \Sur(\P,\Q)$\;
                
            }
        
		\KwRet{$\P$};
	}	
	\caption{Genetic Algorithm}\label{alg:genetic}
\end{algorithm}\DecMargin{1em}

\IncMargin{1em}
\begin{algorithm}
        \SetKwData{P}{P}
        \SetKwData{X}{M}
        \SetKwData{Q}{Q}
        \SetKwData{G}{g}
        \SetKwData{F}{f}
        \SetKwData{N}{n}
        \SetKwData{CM}{cm}
	\SetKwFunction{Init}{randomInitialization}
        \SetKwFunction{Eval}{evaluation}
        \SetKwFunction{Bang}{bang}
        \SetKwFunction{Crunch}{crunch}
        \SetKwFunction{Sel}{selection}
        \SetKwFunction{Var}{variation}
        \SetKwFunction{Sur}{survival}
	\SetKwInOut{Input}{input}
	\SetKwInOut{Output}{output}
        \Input{Population size $\N$, number of generations $\G$}
	\Output{The most fitting solution $\P(1)$}
	\BlankLine		
	
	\Begin{	

    	$\P \leftarrow \Init(\N)$\;
            \For{$i \in [1, \G]$}
            {
                \If{$i \neq 1$}{
                    $\P \leftarrow \Bang(\CM,i)$\;
                }
                $\P \leftarrow \Eval(\P)$\;
                $\CM \leftarrow \Crunch(\P)$\;
                
            }
        
		\KwRet{$\P$};
	}
	
	\caption{Big Bang-Big Crunch Algorithm}\label{alg:BB-BC}
\end{algorithm}\DecMargin{1em}

\subsubsection{Big Bang-Big Crunch Algorithm (BB-BC)}

BB-BC~\cite{erol2006new} is inspired by the evolution of the universe through two phases of explosion and implosion: (i) energy dissipation producing disorder and randomness, and (ii) randomness drawn back into a (different) order. With reference to Algorithm~\ref{alg:BB-BC}, BB-BC creates an initial random population $P$ uniformly spread throughout the search space (the explosion, or big bang), evaluates them, and collects them into their center of mass $cm$ (the implosion, or big crunch). These two phases are repeated throughout the execution, spreading new solutions closer to the center of mass as the number of iterations increases. Re-iterating this procedure leads the center of mass to converge to the optimal solution of the problem. BB-BC is known to outperform GAs in terms of convergence speed; thus, we considered applying it to our problem due to the high run time of the objective function.

\section{Results of the Comparisons}
\label{sec:experiments}

We conducted two main experiments on a computer with 16-core 5.4 GHz CPU and 64 GB RAM using \textit{(i)} three optimization methods (BF, GA, and BB-BC) following the original optimization settings for U-HEx design~\cite{sarac2017design} within the original search space 
and \textit{(ii)} two evolutionary optimization methods (GA and BB-BC) in a wider search space
. These values have been chosen empirically through observations and sensitivity analysis on the feasibility of the solution retrieved (i.e., the solutions are infeasible outside those bounds). Their results will be compared regarding the optimality of the solution and run time. 



Tab.~\ref{tab:param} lists each algorithm's parameters and respective values. 
Most of them did not require any pre-evaluation and were empirically set to a value or a type. For example, the number of generations/iterations was set to $50$ after observing that both algorithms converge earlier (around 35$^{th}$ iteration for GA, and 20$^{th}$ for BB-BC, after which the most fitting solution does not improve more than $0.5$ Nm).  
The probability of performing mutation in GA and the population size require further investigation through preliminary tests to be determined to increase the efficacy of each EC method.

\begin{table}[h!]
\centering
\caption{Parameters of GA and BB-BC}
\label{tab:param}
\resizebox{\columnwidth}{!}{%
\begin{tabular}{|l|c|c|}
\hline
\rowcolor[HTML]{C0C0C0} 
\multicolumn{1}{|c|}{\cellcolor[HTML]{C0C0C0} \textbf{PARAMETER}} & \textbf{PRE. TEST} & \textbf{EXPERIMENT VALUE}                       \\ \hline
\rowcolor[HTML]{FFFFFF} 
Max Num. Generations (GA, BB-BC)                                    &         & $50$                                \\
\rowcolor[HTML]{EFEFEF}  
Population Size (GA, BB-BC)                                               & $150$, $300$ &   $300$                             \\
\rowcolor[HTML]{FFFFFF} 
Selection Type (GA)                                     &            & Binary Tournament~\cite{goldberg1991comparative}                 \\
\rowcolor[HTML]{EFEFEF}
Crossover Type (GA)                                   &              & blx-$\alpha$ ($\alpha = 0.5$)~\cite{eshelman1993real}        \\
\rowcolor[HTML]{FFFFFF} 
Crossover Probability  (GA)                              &            & $1.0$                                \\
\rowcolor[HTML]{EFEFEF} 
Mutation Type (GA)                                         &          & Polynomial~\cite{deb1996combined}                               \\
\rowcolor[HTML]{FFFFFF} 
Mutation Probability (GA)                                         & $0.2$, $0.4$, $0.6$       & $0.2$                          \\
\rowcolor[HTML]{EFEFEF} 
Survival Type (GA)                                      &             & Elitist ($\mu+\lambda$ scheme)~\cite{beyer2002evolution}       \\
\rowcolor[HTML]{FFFFFF} 
Crunch Method (BB-BC)                                      &             & Best Fit~\cite{gencc2010big}\\ 
\rowcolor[HTML]{EFEFEF} 
Constraint Handling (GA, BB-BC)                       &                   & Deb's method~\cite{coello2002theoretical}                               \\ \hline
\end{tabular}}
\end{table}


\subsection{Preliminary Tests on Optimization Methods Parameters}
\label{sec:preliminary}

There is no strict rule for setting the probability of performing mutation of the GA ($PoM$), even though the literature suggests keeping this value low to favor the exploitation of a good solution~\cite{goldberg1987genetic}. We executed the GA ten times for each different value ($3_{PoM} = 0.2$, $0.4$, $0.6$) on a population of 150 solutions 
and observed the respective designs. 
The results of a one-way Analysis of Variation (ANOVA) indicate that there are no statistically significant differences among them ($F(2, 8) = 0.913$, $p = 0.419$, $\eta^2 = 0.092$).

Regarding the population size ($P$), larger values correspond to higher search-space exploration but negatively affect the run time as the number of evaluations increases linearly. 
We executed the GA and the BB-BC independently ten times (each) for different population sizes ($2_{P} = 150$, $300$) and observed their effect on the solutions. 
For GA, our t-test results indicate that population 300 obtains statistically significantly higher forces than 150 ($p = 0.012$) against a higher run time ($p = 0.004$). In contrast, for BB-BC, our t-test results indicate that population 300 obtains a statistically significant difference from 150 in terms of run time ($p = 0.005$) but not the obtained forces ($p = 0.162$). 

Based on these results, we performed the main experiments with a fixed value for ${PoM = 0.2}$ and the population size ${P = 300}$, as also summarized in Table~\ref{tab:param}.

\begin{table}[h]
    \centering
    \caption{Link Lengths Bounds (Experiment 1)}
    \label{tab:bounds1}
    \begin{tabular}{cccccc}
    $\overline{BC}$ & $\overline{CD}$ & $\overline{DE}$ & $\overline{EF}$ & $\overline{FG}$	& $\overline{GH}$ \\ \hline
    $38 \rightarrow 60$ & $10 \rightarrow 30$ & $15 \rightarrow 51$ & $15 \rightarrow 51$ & $20 \rightarrow 56$ & $64 \rightarrow 100$ \\ 
    \end{tabular}
    \vspace{-20pt}
\end{table}

\begin{table}[h]
    \centering
    \caption{Link Lengths Bounds (Experiment 2)}
    \label{tab:bounds2} 
    \begin{tabular}{ccc}
    $\overline{BK}$ &	$\overline{CI}$ & $\overline{EJ}$ \\ \hline
    $20 \rightarrow 50$ & $10 \rightarrow 17$ & $20 \rightarrow 50$ \\ 
    \end{tabular}
\end{table}

\begin{figure}[t!]
    \centering
    \includegraphics[width=0.5 \textwidth]{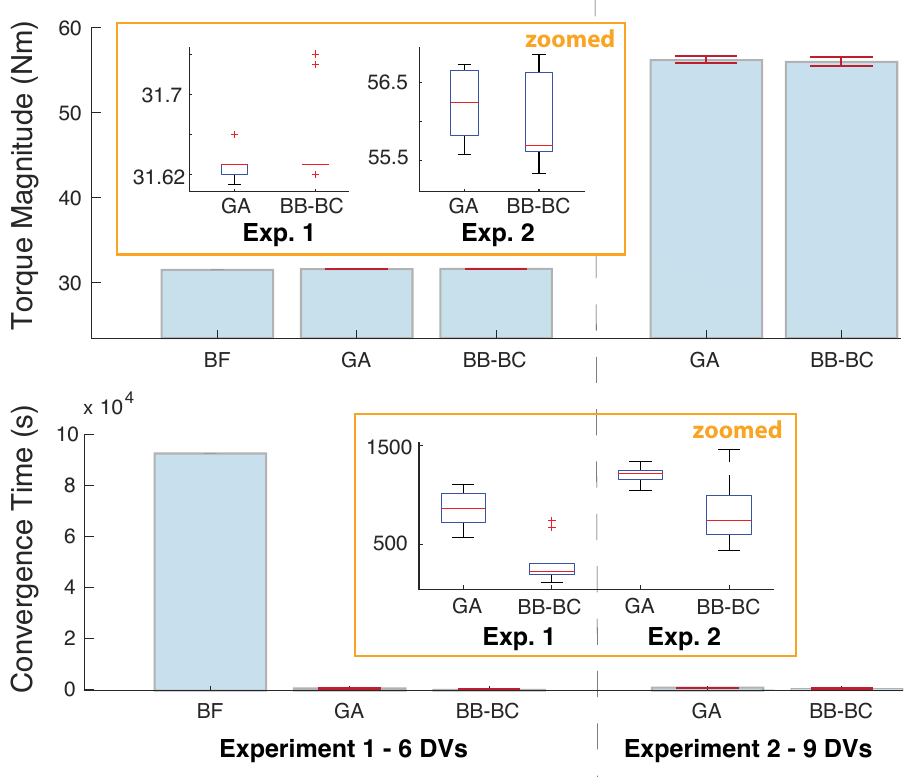}
        
    \caption{Comparison between results of Experiments 1 and 2. Plots include median, interquartile range, and outliers. 
    Details are zoomed in to appreciate variance amongst different conditions.}
    \label{fig:expResults}
        \vspace{-10pt}
\end{figure}

\subsection{Experiment 1: Comparison with Previous Work}
\label{sec:exp1}

We evaluated the impact of EC on U-HEx design by comparing the optimality of their retrieved solutions against the original BF design~\cite{sarac2017design}. For a valid comparison, the search space of the chosen 6 decision variables is kept constant as in our previous work, summarized in Table~\ref{tab:bounds1}.
To compare the run time as well, we re-executed BF. Since BF is expected to provide the same output at every run, we executed it only once, whereas we executed both GA and BB-BC twenty times each. At each execution, we recorded the set of optimized link lengths (i.e., values of decision variable for the most fitting solution) and compared the optimization methods in terms of \textit{(i)} the optimality measure (i.e., torque magnitude) and \textit{(ii)} the run time until convergence (i.e., when the improvement in the most fitting solution's value was smaller than $0.5$ Nm). 

\begin{table}[b!]
    \centering
    \vspace{-10pt}
    \caption{Numerical results of Experiment 1 (6 DVs)}
        \vspace{-5pt}
    \label{tab:exp1}
    \begin{tabular}{l|c|c}
    \multicolumn{1}{c|}{\textbf{}} & \textit{\textbf{Optimality (Nm)}} & \textit{\textbf{Run Time (s)}} \\ \hline
    \textbf{Brute Force} & $31.53 \pm 0.00$ & $94830.49 \pm 0.00$ \\
    \textbf{GA} & $31.63 \pm 0.01$ & $860.51 \pm 168.25$ \\
    \textbf{BB-BC} & $31.65 \pm 0.04$ & $312.50 \pm 204.63$ \\ 
    \end{tabular}
\end{table}

\paragraph{Optimality} Fig.~\ref{fig:expResults} up-left and Table \ref{tab:exp1} show the optimized joint torques for all optimization methods. The findings of one-way ANOVA indicate these methods to be statistically significantly different than each other ($F(2, 8) = 280.515$, $p < 0.001$, $\eta^2 = 0.986$). We then performed a post-hoc analysis with the Bonferroni test: the optimal solution obtained with BF is found to be significantly worse than GA ($p < 0.001$) and than BB-BC ($p < 0.001$) but not between GA and BB-BC ($p = 0.269$). 

\paragraph{Run Time} Fig.~\ref{fig:expResults} down-left and Table~\ref{tab:exp1} show the run time until convergence for all optimization methods. The findings of one-way ANOVA indicate these methods to be statistically significantly different than each other ($F(2, 18) = 17882238.147, p < 0.001, \eta^2 = 1.00$). Further post-hoc (Bonferroni) analysis shows that the run time with BF is significantly higher than GA ($p < 0.001$) and than BB-BC ($p < 0.001$), while the run time until convergence is significantly higher with GA than BB-BC ($p < 0.001$). 



\subsection{Experiment 2: Inclusion of More Decision Variables}
\label{sec:exp2}

The previous experiment shows that EC methods improve the optimum performance measures and the run time compared to a naive BF. With the run time decreasing more than 10 times, we can now include other decision variables, which were constant in the previous experiment. We included three additional decision variables (DVs) to the algorithms: BK, EJ, and CI, with the domains reported in Table~\ref{tab:bounds1}.
To emphasize the impact of this comparison, we performed a two-way ANOVA with factors defined as EC ($2_{EC}$ = GA, BB-BC) and DVs ($2_{DVs}$ = 6 DVs, 9 DVs). 

\begin{table}[h!]
\centering
\caption{Numerical results of Experiment 2 (9 DVs)}
    \vspace{-5pt}
\label{tab:exp2}
\begin{tabular}{l|c|c}
\multicolumn{1}{c|}{\textbf{}} & \textit{\textbf{Optimality (Nm)}} & \textit{\textbf{Run Time (s)}} \\ \hline
\textbf{GA} & $56.20 \pm 0.41$ & $1208.59 \pm 78.31$ \\
\textbf{BB-BC} & $55.99 \pm 0.54$ & $802.50 \pm 298.03$ \\ 
\end{tabular}
\end{table}

 

 \begin{table*}[]
\centering
\caption{Best Design retrieved with Brute Force vs Evolutionary Algorithms}
\label{tab:best_design}
\begin{tabular}{l|cccccc|ccc|c}
 & \multicolumn{9}{c|}{\textit{\textbf{Link Lengths (mm)}}} & \multirow{2}{*}{\textit{\textbf{Magnitude (Nm)}}} \\
\textbf{} & $\overline{BC}$ & $\overline{CD}$ & $\overline{DE}$ & $\overline{EF}$ & $\overline{FG}$	& $\overline{GH}$ & $\overline{BK}$ &	$\overline{CI}$ & $\overline{EJ}$ &  \\ \hline
\textbf{Brute Force} & 58.00 & 10.00 & 15.00 & 51.00 & 56.00 & 100.00 & 35.00 & 16.00 & 37.00 & 31.53 \\
\textbf{Evolutionary} & 60.00 & 10.00 & 15.00 & 51.00 & 56.00 & 91.37 & 48.50 & 10.98 & 36.54 & 56.86
\end{tabular}
\vspace{-10pt}
\end{table*}

 \paragraph{Optimality} Fig.~\ref{fig:expResults} up-right and Tab.~\ref{tab:exp2} show the results of the experiment with 9 DVs in terms of torque magnitudes. We found statistical significance between different DVs ($F(1, 36) = 46655.129$, $p < 0.001$, $\eta^2 = 0.999$), but not between EC ($F(1, 36) = 0.739$, $p = 0.396$, $\eta^2 = 0.02$) or interactions ($F(1, 36) = 1.131$, $p = 0.295$, $\eta^2 = 0.03$).

 \paragraph{Run Time} Fig.~\ref{fig:expResults} up, right and Table \ref{tab:exp2} show the results of the second experiment with 9 DVs in terms of the convergence run time. We found statistical significance for the main factors of DVs ($F(1, 36) = 38.281$, $p < 0.001$, $\eta^2 = 0.515$) and for EC ($F(1, 36) = 49.615$, $p < 0.001$, $\eta^2 = 0.58$), but not between interactions ($F(1, 36) = 1.098$, $p = 0.302$, $\eta^2 = 0.30$). 
 


\section{Discussions}

Our main motivation was to systematically compare EC algorithms to naive optimization methods like BF from the perspective of an engineering design problem -- e.g., the mechanical design of a robotic device (U-HEx). With Experiment 1, we compared a previously implemented method with two EC methods and observed that EC methods are statistically significantly better than BF -- with the given domain and restrictions regarding the optimality of the obtained solution~\cite{sarac2017design}. Similarly, the run time recorded with BF is at least ten times higher than EC methods. Therefore, our first two hypotheses \textbf{(H1)} \textit{EC provides better and more optimal solutions than practical BF} and  \textbf{(H2)} \textit{EC methods provide an optimal solution significantly faster than BF} hold true. 

\begin{figure}[b!]
    \centering
\includegraphics[width=0.9\linewidth]{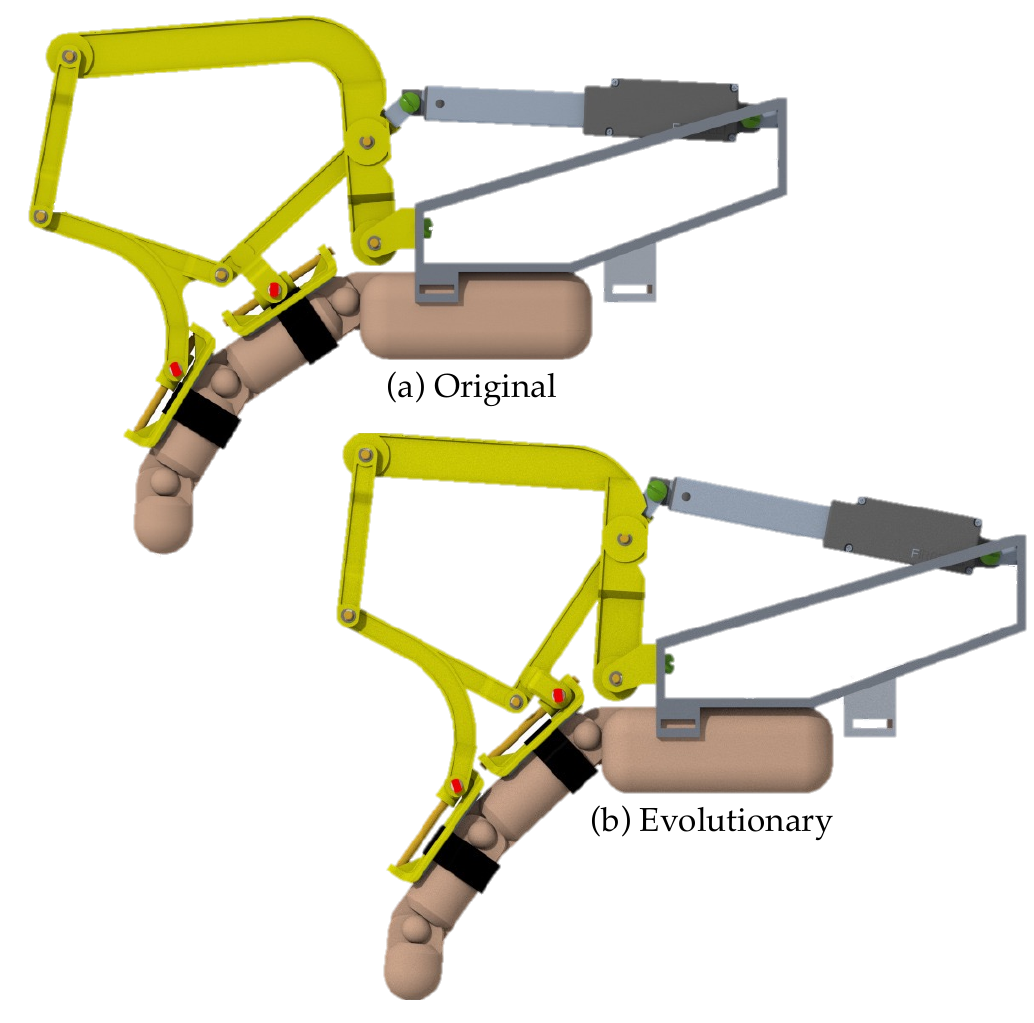}
    \caption{CAD models of U-HEx with the lengths retrieved from (a) BF, similar to the original prototype~\cite{sarac2017design} and (b) EC -- specifically BB-BC.}
    \label{fig:designs}
\end{figure}

The most relevant practical limitation of BF is its run time. Although running BF was computationally possible (with 26 hours run time), three main restrictions were originally made to permit it: discretizing the continuous domain, increasing the step between contiguous values, and reducing the number of decision variables. Based on our results, the first two restrictions hindered more fitting solutions. It is evident that, with the same search space, BF would retrieve \textit{the} optimal solution, outperforming any optimization method in terms of optimality. However, in practice, even with a discrete space and a step of 1 between solutions ($905\CommaPunct219\CommaPunct763$ combinations), we estimate a run time of $4\CommaPunct190$ days
on the same machine 
-- with no guarantee of retrieving the same (sub)optimal EC solution reported in Table~\ref{tab:best_design} and shown in Fig. \ref{fig:designs}, which featured a (pseudo)-continuous search space. In other words, it is possible that the BF could yield solutions that are not statistically different than EC, but EC's superiority in run time would hold true regardless. 

The third restriction was to reduce the number of decision variables from the search space and keep their values constant. Particularly, the sensitivity analysis performed in the original work~\cite{sarac2017design} led designers to remove some decision variables from the problem. Thanks to EC's faster run time, we were able to include three further decision variables in Experiment 2 (Sec.~\ref{sec:exp2}). Our results show that adding more decision variables slightly increases the convergence run time to favor the optimality of the retrieved solution.

Previously, designers of U-HEx used sensitivity analysis to identify the most impactful link lengths as decision variables to make the execution more effective and not waste computation time~\cite{sarac2017design}. Interestingly, Tables~\ref{tab:best_design} and~\ref{tab:bounds1} show that four of these six link lengths ($\overline{CD}$, $\overline{DE}$, $\overline{EF}$, and $\overline{FG}$) remain the same at the mechanical limits imposed by the kinematic chain. Thus, pre-preparing the optimization through search spaces and limited link lengths might require extra attention. Even though with high dimensional spaces, sensitivity analysis can still be useful with EC, it was unnecessary while using EC methods in this study. 



Our third hypothesis \textbf{(H3)} \textit{various EC algorithms might offer different solutions
at different convergence times} was also confirmed. BB-BC converged significantly faster than GA -- which is in line with the claims of the literature~\cite{erol2006new} even though we observed no statistically significant differences between GA and BB-BC in optimality. We also observed that increasing the population size of GA significantly improves the optimality of the retrieved solution -- indicating that larger population sizes might lead to better designs than the one reported in Table~\ref{tab:best_design} and shown in Fig. \ref{fig:designs}. This can also be true for BB-BC, even though we did not find a significant difference in changing the population size (possibly, the observation value was not large enough). However, increasing the population size also increases the overall run time.

Lastly, we would like to emphasize that we do not claim to have proven the superiority of optimization methods against simple and brute enumeration (BF) -- this is a well-known advantage of numerical optimization. 
Instead, our main motivation is to compare different EC methods and provide systematic evidence for the impact of alternative approaches through a hands-on design case of a robotic exoskeleton device (U-HEx). With the enlightenment from our findings, we invite young roboticists, researchers, and designers to be mindful of such alternative methods and choose efficient and effective methods to reach optimality for their design. 
After all, due to this negligence, U-HEx existed in a non-optimal shape for more than a decade.

\section{Conclusions}
\label{sec:conclusion}

In this work, we presented a comparative study to highlight the impact of EC on robotic design. Specifically, we re-optimized the design parameters for U-HEx, an underactuated hand exoskeleton with a numerical, complex kinematic structure -- previously optimized with a naive brute-force method. We showed that EC allowed the device to be further optimized by adding further decision variables. Ultimately, increasing the optimality of the device might actually improve the usability and the efficacy of U-HEx during human interactions during physical rehabilitation therapy for patients with hand disabilities. The code is available for on EVO Lab’s MathWorks File Exchange repository\footnote{\url{www.mathworks.com/matlabcentral/fileexchange/157446}}. 

In the future, we will compare our results with more optimization methods, and expand the study with further objectives to optimize U-HEx (e.g., balancing the forces on each joint to minimize the possibility of hurting the user and reducing the size of the exoskeleton to promote comfort and portability). This investigation will require the implementation of specific multi-objective optimization algorithms with EC and cannot be achieved by BF. We will also investigate the implications of the proposed improvements during the real human-robot interaction by manufacturing both designs and studying the user experience and interaction forces. 






\bibliographystyle{IEEEtran}
\bibliography{references}

\begin{thebibliography}{10}
\providecommand{\url}[1]{#1}
\csname url@samestyle\endcsname
\providecommand{\newblock}{\relax}
\providecommand{\bibinfo}[2]{#2}
\providecommand{\BIBentrySTDinterwordspacing}{\spaceskip=0pt\relax}
\providecommand{\BIBentryALTinterwordstretchfactor}{4}
\providecommand{\BIBentryALTinterwordspacing}{\spaceskip=\fontdimen2\font plus
\BIBentryALTinterwordstretchfactor\fontdimen3\font minus
  \fontdimen4\font\relax}
\providecommand{\BIBforeignlanguage}[2]{{%
\expandafter\ifx\csname l@#1\endcsname\relax
\typeout{** WARNING: IEEEtran.bst: No hyphenation pattern has been}%
\typeout{** loaded for the language `#1'. Using the pattern for}%
\typeout{** the default language instead.}%
\else
\language=\csname l@#1\endcsname
\fi
#2}}
\providecommand{\BIBdecl}{\relax}
\BIBdecl

\bibitem{poliero2020applicability}
T.~Poliero, M.~Lazzaroni, S.~Toxiri, C.~Di~Natali, D.~G. Caldwell, and
  J.~Ortiz, ``Applicability of an active back-support exoskeleton to carrying
  activities,'' \emph{Frontiers in Robotics and AI}, vol.~7, p. 579963, 2020.

\bibitem{mauri2019mechanical}
A.~Mauri, J.~Lettori, G.~Fusi, D.~Fausti, M.~Mor, F.~Braghin, G.~Legnani, and
  L.~Roveda, ``Mechanical and control design of an industrial exoskeleton for
  advanced human empowering in heavy parts manipulation tasks,''
  \emph{Robotics}, vol.~8, no.~3, p.~65, 2019.

\bibitem{butzer2021fully}
T.~B{\"u}tzer, O.~Lambercy, J.~Arata, and R.~Gassert, ``Fully wearable actuated
  soft exoskeleton for grasping assistance in everyday activities,'' \emph{Soft
  robotics}, vol.~8, no.~2, pp. 128--143, 2021.

\bibitem{koyama2002multi}
T.~Koyama, I.~Yamano, K.~Takemura, and T.~Maeno, ``Multi-fingered exoskeleton
  haptic device using passive force feedback for dexterous teleoperation,'' in
  \emph{IEEE/RSJ International Conference on Intelligent Robots and Systems},
  vol.~3, 2002, pp. 2905--2910.

\bibitem{stroppa2017robot}
F.~Stroppa, C.~Loconsole, S.~Marcheschi, and A.~Frisoli, ``A robot-assisted
  neuro-rehabilitation system for post-stroke patients’ motor skill
  evaluation with alex exoskeleton,'' in \emph{Proceedings of the International
  Conference on NeuroRehabilitation (ICNR)}, 2017, pp. 501--505.

\bibitem{marcheschi2011body}
S.~Marcheschi, F.~Salsedo, M.~Fontana, and M.~Bergamasco, ``Body extender:
  {W}hole body exoskeleton for human power augmentation,'' in \emph{IEEE
  International Conference on Robotics and Automation (ICRA)}, 2011, pp.
  611--616.

\bibitem{gijbels2011armeo}
D.~Gijbels, I.~Lamers, L.~Kerkhofs, G.~Alders, E.~Knippenberg, and P.~Feys,
  ``The armeo spring as training tool to improve upper limb functionality in
  multiple sclerosis: {A} pilot study,'' \emph{Journal of Neuroengineering and
  Rehabilitation}, vol.~8, pp. 1--8, 2011.

\bibitem{zoss2006biomechanical}
A.~B. Zoss, H.~Kazerooni, and A.~Chu, ``Biomechanical design of the berkeley
  lower extremity exoskeleton (bleex),'' \emph{IEEE/ASME Transactions on
  Mechatronics}, vol.~11, no.~2, pp. 128--138, 2006.

\bibitem{buongiorno2018wres}
D.~Buongiorno, E.~Sotgiu, D.~Leonardis, S.~Marcheschi, M.~Solazzi, and
  A.~Frisoli, ``Wres: {A} novel 3 {DoF} {WRist ExoSkeleton} with tendon-driven
  differential transmission for neuro-rehabilitation and teleoperation,''
  \emph{IEEE Robotics and Automation Letters (RA-L)}, vol.~3, no.~3, pp.
  2152--2159, 2018.

\bibitem{sarac2017design}
M.~Sarac, M.~Solazzi, E.~Sotgiu, M.~Bergamasco, and A.~Frisoli, ``Design and
  kinematic optimization of a novel underactuated robotic hand exoskeleton,''
  \emph{Meccanica}, vol.~52, pp. 749--761, 2017.

\bibitem{sarac2019design}
M.~Sarac, M.~Solazzi, and A.~Frisoli, ``Design requirements of generic hand
  exoskeletons and survey of hand exoskeletons for rehabilitation, assistive,
  or haptic use,'' \emph{IEEE Transactions on Haptics (ToH)}, vol.~12, no.~4,
  pp. 400--413, 2019.

\bibitem{stroppa2023optimizing}
F.~Stroppa, A.~Soylemez, H.~T. Yuksel, B.~Akbas, and M.~Sarac, ``Optimizing
  exoskeleton design with evolutionary computation: An intensive survey,''
  \emph{Robotics}, vol.~12, no.~4, p. 106, 2023.

\bibitem{sioshansi2017optimization}
R.~Sioshansi and A.~J. Conejo, \emph{Optimization in Engineering}.\hskip 1em
  plus 0.5em minus 0.4em\relax Cham: Springer International Publishing, 2017,
  vol. 120.

\bibitem{statnikov2012multicriteria}
R.~B. Statnikov and J.~B. Matusov, \emph{Multicriteria Optimization and
  Engineering}.\hskip 1em plus 0.5em minus 0.4em\relax Springer Science and
  Business Media, 2012.

\bibitem{andersson2000survey}
J.~Andersson, ``A survey of multiobjective optimization in engineering
  design,'' \emph{Department of Mechanical Engineering, Linktjping University.
  Sweden}, 2000.

\bibitem{bonnans2006numerical}
J.-F. Bonnans, J.~C. Gilbert, C.~Lemar{\'e}chal, and C.~A. Sagastiz{\'a}bal,
  \emph{Numerical Optimization: Theoretical and Practical Aspects}.\hskip 1em
  plus 0.5em minus 0.4em\relax Springer Science and Business Media, 2006.

\bibitem{dumitrescu2000evolutionary}
D.~Dumitrescu, B.~Lazzerini, L.~C. Jain, and A.~Dumitrescu, \emph{Evolutionary
  Computation}.\hskip 1em plus 0.5em minus 0.4em\relax CRC press, 2000.

\bibitem{laliberte2002underactuation}
T.~Laliberte, L.~Birglen, and C.~Gosselin, ``Underactuation in robotic grasping
  hands,'' \emph{Machine Intelligence and Robotic Control}, vol.~4, no.~3, pp.
  1--11, 2002.

\bibitem{buryanov2010proportions}
A.~Buryanov and V.~Kotiuk, ``Proportions of hand segments,'' \emph{Int. J.
  Morphol}, pp. 755--758, 2010.

\bibitem{norde2000characterizing}
H.~Norde, F.~Patrone, and S.~Tijs, ``Characterizing properties of approximate
  solutions for optimization problems,'' \emph{Mathematical Social Sciences},
  vol.~40, no.~3, pp. 297--311, 2000.

\bibitem{nomaguchi2016robust}
Y.~Nomaguchi, K.~Kawakami, K.~Fujita, Y.~Kishita, K.~Hara, and M.~Uwasu,
  ``Robust design of system of systems using uncertainty assessment based on
  lattice point approach: Case study of distributed generation system design in
  a japanese dormitory town,'' \emph{International Journal of Automation
  Technology}, vol.~10, no.~5, pp. 678--689, 2016.

\bibitem{dizangian2015reliability}
B.~Dizangian and M.~Ghasemi, ``Reliability-based design optimization of complex
  functions using self-adaptive particle swarm optimization method,''
  \emph{International Journal of Optimization in Civil Engineering}, vol.~5,
  no.~2, pp. 151--165, 2015.

\bibitem{goldberg1989genetic}
D.~E. Goldberg, \emph{Genetic Algorithms in Search, Optimization, and Machine
  Learning}.\hskip 1em plus 0.5em minus 0.4em\relax Addison-Wesley, 1989.

\bibitem{goldberg1990real}
------, \emph{Real-Coded Genetic Algorithms, Virtual Alphabets and Blocking},
  1991, vol.~5, no.~2.

\bibitem{erol2006new}
O.~K. Erol and I.~Eksin, ``A new optimization method: Big {B}ang--{B}ig
  {C}runch,'' \emph{Advances in Engineering Software}, vol.~37, no.~2, pp.
  106--111, 2006.

\bibitem{goldberg1991comparative}
D.~E. Goldberg and K.~Deb, ``A comparative analysis of selection schemes used
  in genetic algorithms,'' in \emph{Foundations of genetic algorithms}.\hskip
  1em plus 0.5em minus 0.4em\relax Elsevier, 1991, vol.~1, pp. 69--93.

\bibitem{eshelman1993real}
L.~J. Eshelman and J.~D. Schaffer, ``Real-coded genetic algorithms and
  interval-schemata,'' in \emph{Foundations of genetic algorithms}.\hskip 1em
  plus 0.5em minus 0.4em\relax Elsevier, 1993, vol.~2, pp. 187--202.

\bibitem{deb1996combined}
K.~Deb, M.~Goyal \emph{et~al.}, ``A combined genetic adaptive search (geneas)
  for engineering design,'' \emph{Computer Science and informatics}, vol.~26,
  pp. 30--45, 1996.

\bibitem{beyer2002evolution}
H.-G. Beyer and H.-P. Schwefel, ``Evolution strategies--a comprehensive
  introduction,'' \emph{Natural computing}, vol.~1, pp. 3--52, 2002.

\bibitem{gencc2010big}
H.~M. Gen{\c{c}}, I.~Eksin, and O.~K. Erol, ``Big bang-big crunch optimization
  algorithm hybridized with local directional moves and application to target
  motion analysis problem,'' in \emph{2010 IEEE International Conference on
  Systems, Man and Cybernetics}.\hskip 1em plus 0.5em minus 0.4em\relax IEEE,
  2010, pp. 881--887.

\bibitem{coello2002theoretical}
C.~A.~C. Coello, ``Theoretical and numerical constraint-handling techniques
  used with evolutionary algorithms: A survey of the state of the art,''
  \emph{Computer Methods in Applied Mechanics and Engineering}, vol. 191, no.
  11-12, pp. 1245--1287, 2002.

\bibitem{goldberg1987genetic}
D.~E. Goldberg, J.~Richardson \emph{et~al.}, ``Genetic algorithms with sharing
  for multimodal function optimization,'' in \emph{Genetic Algorithms and Their
  Applications: Proceedings of the Second International Conference on Genetic
  Algorithms}, vol. 4149, 1987.

\end{thebibliography}


\end{document}